**Title Page (with Author Details)**

# Three-dimensional Damage Visualization of Civil Structures via Gaussian Splatting-enabled Digital Twins

Shuo Wang[a†], Shuo Wang[b†], Xin Nie[a], Yasutaka Narazaki[b*], Thomas Matiki[c], Billie F. Spencer Jr.[c]

[a]Key Laboratory of Civil Engineering Safety and Durability of China Education Ministry, Department of Civil Engineering, Tsinghua University, Beijing, 100084, China.

[b]Zhejiang University-University of Illinois Urbana-Champaign Institute, Zhejiang University, Haining, 314400, China.

[c]Department of Civil and Environmental Engineering, University of Illinois Urbana-Champaign, Urbana, 61801, U.S.A.

[†]Equal contribution.

*Corresponding author. E-mail: narazaki@intl.zju.edu.cn

Contributing authors. E-mail: shuow2@mail.tsinghua.edu.cn, shuo.24@intl.zju.edu.cn, xinnie@tsinghua.edu.cn, tmatiki2@illinois.edu, bfs@illinois.edu

**Manuscript (without Author Details)**

# Three-dimensional Damage Visualization of Civil Structures via Gaussian Splatting-enabled Digital Twins

**Abstract:**

Recent advancements in civil infrastructure inspections underscore the need for precise three-dimensional (3D) damage visualization on digital twins, transcending traditional 2D image-based damage identifications. Compared to conventional photogrammetric 3D reconstruction techniques, modern approaches such as Neural Radiance Field (NeRF) and Gaussian Splatting (GS) excel in scene representation, rendering quality, and handling featureless regions. Among them, GS stands out for its efficiency, leveraging discrete anisotropic 3D Gaussians to represent radiance fields, unlike NeRF's continuous implicit model.

This study introduces a GS-enabled digital twin method tailored for effective 3D damage visualization. The method's key contributions include: 1) utilizing GS-based 3D reconstruction to visualize 2D damage segmentation results while reducing segmentation errors; 2) developing a multi-scale reconstruction strategy to balance efficiency and damage detail; 3) enabling digital twin updates as damage evolves over time. Demonstrated on an open-source synthetic dataset for post-earthquake inspections, the proposed approach offers a promising solution for comprehensive 3D damage visualization in civil infrastructure digital twins.

**Keywords:** Gaussian Splatting, digital twin, 3D damage visualization

## 1. Introduction

Inspections play a pivotal role in ensuring the safety and longevity of civil infrastructure. Post-event inspections, such as those conducted after earthquakes or hurricanes [1] [2] [3] , are crucial for quickly assessing structural integrity and identifying immediate hazards, guiding emergency repairs. Routine inspections [4] [5] [6] , on the other hand, enable the early detection of gradual wear and tear, allowing for timely maintenance to prevent major failures. Recent research regarding structural inspections actively investigates damage detection and segmentation from 2D images, leveraging computer vision techniques and machine learning algorithms [7] [8] . For example, Mondal et al. [9] used Region-based convolutional neural network (Faster-RCNN) to detect four different damage types, namely, surface crack, spalling, severe damage with exposed rebars and severely buckled rebars on reinforced concrete buildings damaged under earthquakes. Yao et al. [10] employed DeepLabV3+ neural network for pixelwise crack identification for high-rise buildings. Huang et al. [11] utilized an improved framework based on Mask R-CNN to optimize multiclass damage detection across various bridge components under natural backgrounds. Li et al. [12] proposed a hybrid U-shaped model based on convolutional neural networks (CNN) and Vision Transformer (ViT) for semantic segmentation of dam cracks. However, understanding the implications of damage for the overall condition of individual structures remains challenging without contextual information [13] [14] , such as the regions where damage is present.

Precise three-dimensional (3D) damage visualization on digital twins is crucial for accurately assessing the implications of identified structural damage [2] [15] . For effective digital twin applications for 3D damage visualization, several key properties are essential. First, multi-scale representational flexibility is critical to balance computational efficiency and damage detail, allowing the digital twin to adaptively prioritize high-fidelity reconstruction in critical damaged areas while maintaining broader scene scalability. Second, the digital twin should possess error-mitigation capabilities, enabling it to reconcile and refine 2D damage segmentation results through 3D reconstruction, thereby reducing inaccuracies inherent in single-view analyses. Lastly, the digital twin must exhibit easy updateability, supporting integration of new damage data as it evolves over time, ensuring the model remains a faithful, temporally consistent reflection of the physical structure's condition.

A prevailing methodology for 3D damage visualization involves a two-step process: reconstructing the 3D scene using photogrammetric techniques and projecting 2D image-derived damage onto the reconstructed model [16] . Traditional photogrammetric 3D reconstruction pipelines, exemplified by the SfM-MVS-Poisson workflow, typically encompass three sequential stages [17] : (1) sparse point cloud generation via Structure from Motion (SfM), which estimates camera parameters and reconstructs feature correspondences; (2) dense point cloud refinement using Multi-View Stereo (MVS), which leverages geometric constraints to densify the sparse reconstruction; and (3) surface mesh creation through Poisson surface reconstruction, which generates a watertight triangular mesh from the dense point cloud. This image-based 3D reconstruction technique gained significant popularity after the publication of landmark studies such as *"Photo Tourism"* [18] and *"Building Rome in a Day"* [19] . Consequently, off-the-shelf commercial software for 3D scene reconstruction has emerged, including RealityScan [20], Agisoft Metashape [21] , and Pix4D [22] . 3D reconstruction technology has since evolved in the field of structural inspection, with notable applications including [23] [24] [25] , etc. Damage projection onto the mesh is commonly realized through ray casting, where rays originating from damaged pixels in 2D images are traced through the camera's optical center to intersect the 3D model. The problem is simplified when the damaged surface in the 3D model is planar, as this reduces the task to a plane-to-plane mapping. Relevant research includes Liu et al. [26] , who projected detected cracks onto the planar surfaces of a concrete reaction wall and a concrete flange, and Chaiyasarn et al. [27] , who projected detected cracks onto a planar surface of the 3D model of a concrete bridge footing structure. Notable applications involving non-planar damaged surfaces include Liu et al. [28] , who projected UAV-detected cracks onto 3D bridge pier reconstructions, and Zhao et al. [29] , who projected identified concrete dam damage onto corresponding 3D models after damage identification in 2D images via YOLOv5.

However, these "photogrammetric 3D reconstruction followed by ray-casting damage projection" methodologies face critical limitations. First, their heavy reliance on feature point correspondences [17] leads to failures in texture-less or repetitive regions. Second, when multiple images capture the same damage, determining which image's damage annotation to cast becomes problematic—a challenge further exacerbated by potential contradictions arising from damage identification errors in overlapping images. Third, the ray-casting process for damage projection, which is by essence a collision detection problem [28] , introduces substantial computational overhead and imposes strict requirements on the precision of camera poses for the images undergoing ray-casting. Finally, poor novel view synthesis capability impedes convenient damage tracking during new inspections when new perspectives are involved. Collectively, these constraints undermine the accuracy, flexibility, and adaptability needed for comprehensive 3D damage visualization.

Modern alternative 3D reconstruction techniques, such as Neural Radiance Fields (NeRF) [30] and Gaussian Splatting (GS) [31] , differ significantly from traditional photogrammetric 3D reconstructions in that they rely on training with neural networks or parametric models to learn scene representations, rather than purely geometric algorithms for feature matching and surface reconstruction. NeRF employs Multi-Layer Perceptron (MLP) to implicitly model the 3D scene as a continuous function of color and density. This allows it to reconstruct intricate details, including fine textures and complex lighting effects, with high fidelity. However, NeRF's reliance on continuous representations leads to challenges in training efficiency. While subsequent NeRF variants such as Mip-NeRF [32] , EfficientNeRF [33] , Instant-NGP [34] , Zip-NeRF [35] , and TensoRF [36] enhanced computational efficiency, they retained the continuous implicit representation paradigm of NeRF, which inherently limits direct geometric manipulation. In contrast, GS [31] explicitly represents the scene using a set of 3D Gaussian distributions parameterized by position, covariance, color, and opacity. This explicit representation enables faster convergence during training and supports direct manipulation of geometric primitives, making it particularly well-suited for applications requiring precise 3D damage visualization. By striking a balance between reconstruction accuracy, geometric controllability, and training cost, GS offers a practical solution for scenarios demanding both high-fidelity 3D modeling and efficient data processing.

In this research, a novel GS-empowered digital twin method is developed to facilitate 3D damage visualization, aiming to overcome the limitations of traditional approaches. The method's key contributions are multi-faceted. First, unlike the conventional two-step process of "photogrammetric 3D reconstruction followed by ray-casting damage projection", the proposed method integrates damage visualization directly into the GS-based 3D reconstruction workflow by incorporating 2D damage segmentation results as ground truth into the loss function. This is accomplished by comparing images rendered from the 3D GS model against ground truth segmentation masks, enabling the model to learn damage distributions during the multi-view consistency optimization process. By enforcing multi-view consistency, this approach effectively mitigates the segmentation errors that commonly affect single-view analyses. Second, the method adopts a hierarchical GS-based reconstruction strategy started from generating a coarse-grained model from low-resolution images to rapidly establish an overview of the structure, thereby conserving computational resources. For regions with suspected or confirmed damage, targeted fine-tuning or retraining is conducted using high-resolution images to achieve high-fidelity, detailed reconstruction. This ensures critical damage features are accurately captured while obviating the need for unnecessary global refinement. Third, leveraging the novel-view synthesis capability inherent to GS, the method enables convenient updates of the 3D model to reflect damage progression. Specifically, newly collected images are compared with views rendered from the historical GS model (captured at previous time points) from identical perspectives, allowing the identification of newly developed damage in 2D. Subsequently, the newly detected damage is integrated into the 3D model through targeted fine-tuning or retraining—with distinct annotations to differentiate it from pre-existing damage. In this way, the proposed approach not only ensures the 3D model remains aligned with real-world structural changes but also preserves clear temporal distinction between damage phases—all without the need for resource-heavy full re-reconstruction.

The three contributions outlined above position the proposed GS-based method as a viable digital twin solution for 3D damage visualization of civil structures. The proposed approach does not seek to overhaul the core GS pipeline; instead, it focuses on modifying components such as the loss function and training strategy to adapt GS for 3D damage visualization tasks. The effectiveness of the proposed approach is then demonstrated on an open-source synthetic dataset specifically designed for post-earthquake inspections. Results show that the GS-enabled digital twin approach accurately visualizes 3D damage, achieving a balance between computational efficiency and damage fidelity while enabling seamless updates. As such, the method offers a robust and scalable solution for comprehensive 3D damage visualization in civil infrastructure digital twins, paving the way for advanced structural health assessment and maintenance strategies. Python scripts developed for this research, which build upon the original GS framework, are publicly available in the GitHub repository hosted at https://github.com/shuow2/Gaussian-Splatting-enabled-digital-twins-for-3D-damage-visualization-of-civil-structures.

## 2. Methodologyss

Section 2 describes the methodology, beginning with the principles of the GS framework for 3D reconstruction (Section 2.1). It then introduces the adaptation of GS for 3D damage visualization (Section 2.2), demonstrating how damage segmentation masks are integrated into the loss function during optimization to present damage segmentation results within their spatial structural context. Building on this, Section 2.3 proposes a hierarchical strategy to save computational costs: low-resolution images are first used to rapidly establish a coarse structural overview, followed by targeted fine-tuning or retraining of damage-related Gaussians using high-resolution images with segmentation masks, ensuring detailed damage capture without global refinement. Section 2.4 leverages GS's novel-view synthesis capacity to facilitate digital twin updates, enabling cross-temporal comparison of damage progression via alignment between model-rendered views and newly collected images, and achieving localized model updates without the need for full re-reconstruction. Finally, Section 2.5 provides a summary of the proposed methodology, while highlighting its key contributions and core advantages.

*2.1 Gaussian Splatting-based 3D reconstruction*

3D GS is a state-of-the-art 3D scene representation and rendering technique that explicitly models scenes using a set of anisotropic 3D Gaussian primitives. Each 3D gaussian primitive is characterized by its center position $\mu \in \mathbb{R}^3$, and a covariance matrix $\Sigma \in \mathbb{R}^{3 \times 3}$ describing its scale $S$ and orientation $R$ in 3D space, expressed as:

$$\Sigma = RSS^T R^T \tag{1}$$

allowing the adaptive fitting of complex surface geometries, from smooth curves to sharp edges. The 3D Gaussian distribution of $X \in \mathbb{R}^3$:

$$N(X; \mu, \Sigma) = \frac{1}{(2\pi)^{3/2} \sqrt{\det(\Sigma)}} \exp(-\frac{1}{2}(X - \mu)^T \Sigma^{-1}(X - \mu)) \tag{2}$$

controls the spatial extent of the primitive, with higher density near the center and exponential decay toward the edges. GS handles appearance through color and opacity parameters. The color $c \in \mathbb{R}^3$ is typically stored as RGB values and often enhanced with spherical harmonics (SH) to model view-dependent effects (e.g., specular highlights, shadowing). The opacity $\alpha \in [0, 1]$ is tied to the Gaussian's probability density, ensuring smooth transitions between overlapping primitives.

The rendering pipeline involves projecting 3D Gaussian ellipsoids into 2D image space as ellipses, a step specifically referred to as "splatting". Given the 3D covariance matrix $\Sigma$ and the viewing transformation matrix $W$, the 2D covariance matrix $\Sigma'$ characterizing the projected 2D Gaussian is computed through:

$$\Sigma' = JW\Sigma W^T J^T \tag{3}$$

where $J$ is the Jacobian of the affine approximation of the projective transformation. Given the position of a pixel $x$ on the 2D image, its distance to all overlapping Gaussians can be computed through the viewing transformation matrix $W$, forming a sorted list of $N$ Gaussians. Then, $\alpha$-blending is adopted to compute the final color of this pixel as:

$$\sum_{n=1}^{N} c_n \alpha'_n \prod_{j=1}^{n-1} \left(1 - \alpha'_j\right) \tag{4}$$

with

$$\alpha' = \alpha \times \exp(-\frac{1}{2}(x - \mu')^T \Sigma'^{-1}(x - \mu')) \tag{5}$$

where $\mu'$ represents the center position of the projected 2D Gaussian. 3D GS enhances rendering efficiency via GPU-accelerated parallel processing of projected 2D ellipses, leveraging tile-based rasterization and early depth testing to cull occluded primitives and minimize redundant computations.

GS reconstructs 3D scenes from multi-view images via an optimization-based methodology, as outlined in Figure 1. Similar to photogrammetry-based approaches, GS-based 3D reconstruction employs Structure from Motion (SfM) as its foundational initial step. Specifically, SfM outputs camera poses and a sparse point cloud—both inherently up-to-scale, meaning they are defined by relative rather than absolute physical dimensions. The camera poses facilitate the rendering pipeline, while the sparse point cloud provides the initial positional and color attributes for the Gaussian primitives. These primitives are typically initialized with a uniform spherical shape and consistent opacity values. The pipeline then enters an iterative refinement phase, where the positions, shapes, colors, and opacities of the 3D Gaussian primitives are progressively adjusted to minimize the discrepancy between rendered views and the original multi-view images. The loss function is a carefully designed combination that typically includes pixel-wise color differences (e.g., L1 loss for penalizing absolute pixel intensity discrepancies) and Structural Similarity (SSIM) to ensure both color and structural fidelity. SSIM evaluates three key components: luminance, contrast, and structure, and aligns more closely with human visual perception compared to traditional pixel-to-pixel intensity error measures. Regularization terms are also added to prevent overfitting and enforce physically plausible Gaussian parameters, such as bounding their size and shape regularity. To perform the optimization, gradients are backpropagated through the entire rendering pipeline, which enables the model to learn fine-grained details, including texture patterns and lighting effects, directly from the input data. Since the optimal number of 3D Gaussians for representing a scene is unknown beforehand, density control strategies are employed. These strategies, such as clone, split, and prune operations, are crucial for adapting the Gaussian density. Cloning duplicates Gaussians in areas that require more detail, splitting divides a Gaussian into smaller ones to capture finer features, and pruning removes redundant or less-significant Gaussians, thus optimizing the scene representation for both accuracy and computational efficiency.

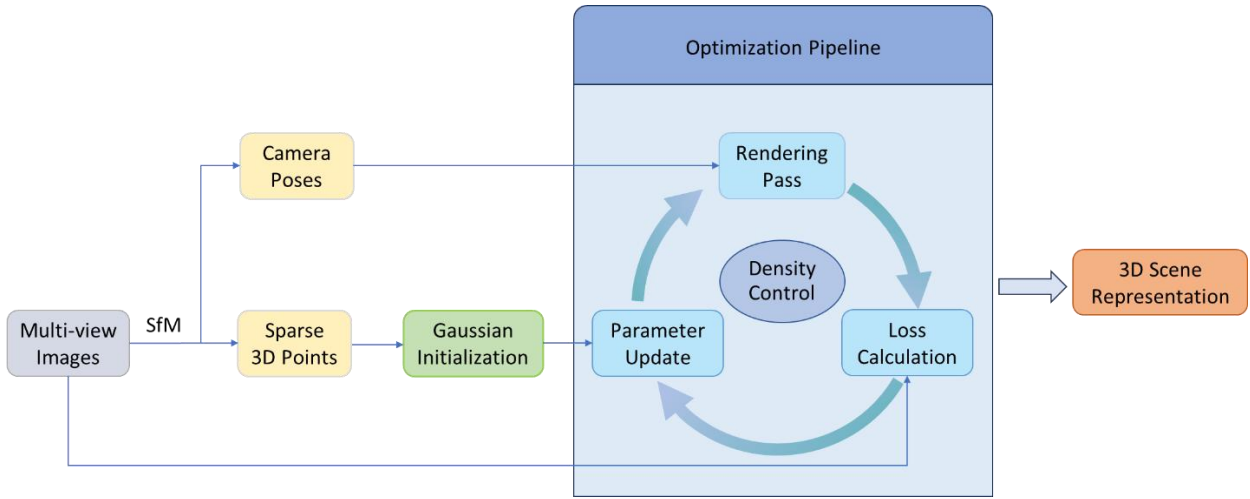

Figure 1. Procedure of GS-based 3D reconstruction

For more in-depth information on 3D GS, readers are recommended to consult the original research paper [31] and comprehensive review papers such as [37] [38] , which delve into its technical framework, optimization strategies, and practical applications.

### 2.2 Gaussian Splatting-enabled 3D damage visualization

Beyond visualizing 3D scenes from field-collected images, GS enables contextual visualization of damage segmentation results by integrating damage visualization directly into the GS-based 3D reconstruction workflow. Specifically, damage segmentation masks (e.g., for concrete cracks or spalling) from individual images are overlaid on their corresponding source images and treated as ground truth; the loss function in the GS optimization pipeline then compares the model's rendered outputs not to the original images, but to these damage-segmented images. Notably, using damage-segmented images directly for traditional photogrammetric 3D reconstruction is infeasible: damage regions—typically annotated with a uniform color—lack discernible features required for dense matching.

GS uniquely mitigates single-view segmentation errors through multi-view consistency optimization. When 2D damage masks from different views contain contradictory errors (e.g., over-segmentation in one view vs. under-segmentation in another), the iterative Gaussian refinement process acts like a "majority vote", suppressing inconsistencies by prioritizing overlapping, consensus-driven regions across views. The smooth, probabilistic nature of Gaussian distributions further softens boundary inaccuracies, allowing the 3D model to converge on a coherent, error-robust representation of damage that aligns with multi-view observations.

The procedure of GS-enabled 3D damage visualization, as outlined in Figure 2, closely follows the foundational workflow of standard GS-based 3D reconstruction. In the initial SfM step, original, unaltered images are utilized to estimate camera poses with high precision, as the rich visual features (e.g., edges, textures, and key points) in these images are indispensable for robust geometric calibration. A critical divergence arises during the optimization phase: instead of comparing rendered outputs against the original images, the loss function is computed using damage segmentation masked images. These masks, generated via advanced deep learning techniques such as DeepLabV3+ [39] , delineate damaged regions at pixel-level accuracy. As a result, the final 3D model not only captures the structural geometry but also contextualizes damage within its spatial environment, enabling detailed, interpretable visualizations that are invaluable for applications like infrastructure inspection and disaster assessment.

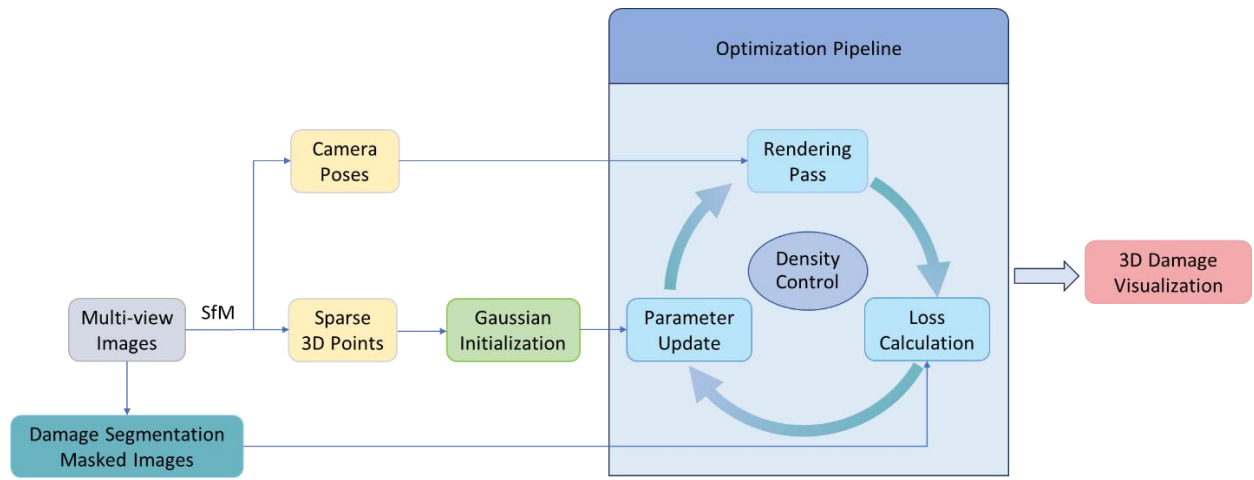

Figure 2. Procedure of GS-enabled 3D damage visualization

## 2.3 Hierarchical 3D damage visualization via Gaussian Splatting

To address the computational challenges of reconstructing large-scale structures while ensuring detailed damage visualization, the method employs a hierarchical reconstruction strategy built on the GS framework. By leveraging the complementary strengths of low- and high-resolution imagery, this approach takes into account both efficiency and accuracy: it first generates a coarse-grained model from low-resolution damage-segmented images to rapidly establish structural overview and conserve computation resources, then applies targeted fine-tuning or retraining to damage regions using high-resolution damage-segmented images. GS's adaptive density control and per-Gaussian parameter refinement allow seamless transitions between coarse global geometry and fine-grained damage details, avoiding unnecessary global refinement while ensuring that critical features are captured with precision.

The procedure of the proposed hierarchical 3D damage visualization strategy is outlined in Figure 3. First, original high-resolution images are used in the SfM step to estimate camera poses with high precision. Subsequently, a baseline reconstruction is generated via the standard GS-enabled 3D damage visualization workflow (summarized in Section 2.2) using low-resolution images overlaid with damage segmentation masks. These masks are assigned a user-defined color to differentiate damage regions from the structural background. Next, 3D Gaussians associated with damage are selected via color-based filtering and projection-based validation, with details illustrated using an example in Section 3. Neighboring Gaussians are also included to capture contextual structural details, and a secondary mask is generated for each view by expanding the convex hull of the 2D projections of these damage-associated and neighboring Gaussians. In the subsequent refinement phase, the loss function between each target image and its corresponding rendered image is computed exclusively within this new mask, thereby confining adjustments to the damage regions.

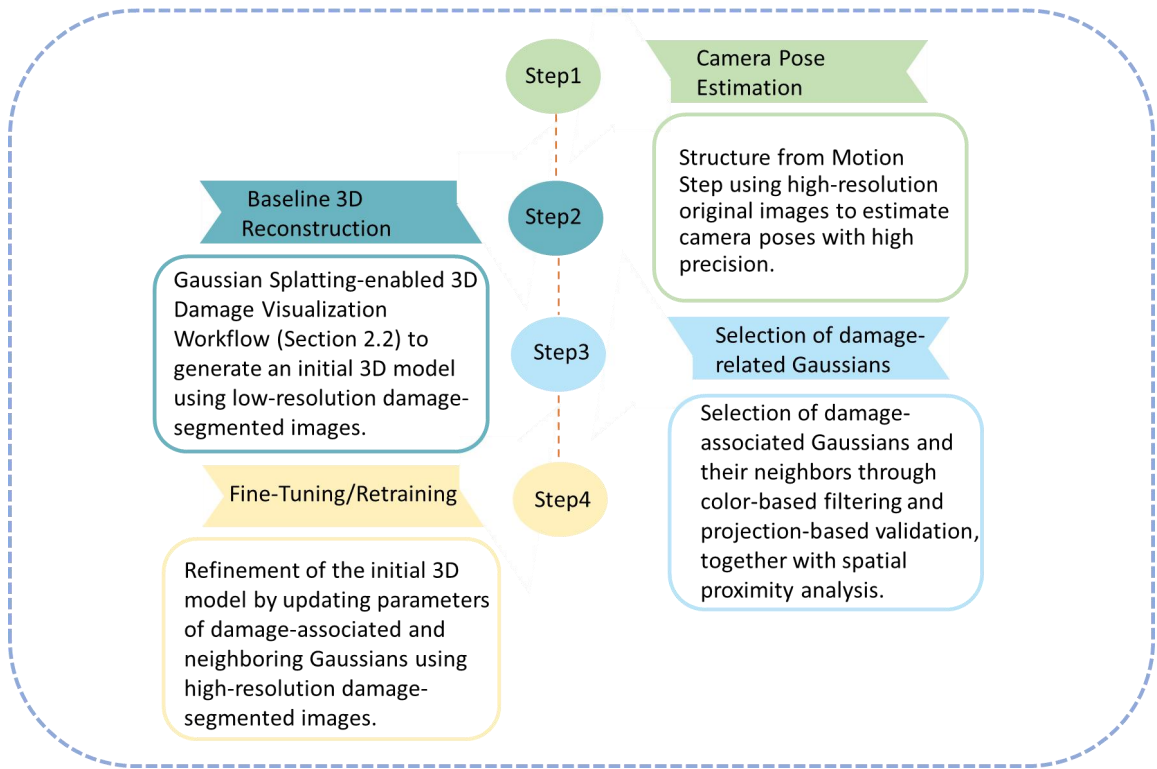

Figure 3. Procedure of hierarchical 3D damage visualization via GS

Subsequently, the selected damage-associated 3D Gaussians and their neighboring Gaussians undergo fine-tuning or retraining using high-resolution images with damage segmentation masks. During this process, non-damage-related Gaussians are frozen to avoid unnecessary parameter updates and conserve computational resources; only the views containing damage are included in the optimization, reducing the number of images processed and focusing efforts on relevant data; the loss is calculated within the expanded convex hull region, limiting adjustments to the specific area of interest and minimizing redundant computations across the entire scene. For fine-tuning, parameters such as position, color, opacity, scaling, and rotation of existing Gaussians are updated without altering their count, while retraining employs densification strategies like clone, split, and prune to dynamically adjust the number of Gaussians for better fit to high-resolution damage-segmented images. The outcome of this hierarchical 3D damage visualization procedure is a 3D model that integrates a coarse-grained overall structure for broad structural context with fine-resolution damage detail in targeted regions, delivering a multi-scale representation optimized for both computational efficiency and visual accuracy.

*2.4 Gaussian Splatting-enabled digital twin updating for damage progression visualization*

Leveraging GS's ability to render high-fidelity views from any calibrated perspective, the method enables precise tracking of damage progression in digital twins. As summarized in Figure 4, the process begins by calibrating newly collected images to determine their camera poses, ensuring alignment with the existing GS model's coordinate system. Using these calibrated perspectives, synthetic views—complete with high-resolution damage masks—are rendered from the pre-existing GS model. Concurrently, the newly collected images undergo damage segmentation to generate their own damage masks. Direct pixel-wise comparison between these new masks and the model-rendered masks then highlights regions where new damage has emerged. Notably, traditional 3D reconstruction methods lack this novel-view synthesis capability, which undermines the practicality of cross-temporal comparisons. On one hand, camera poses between old and new surveys are not necessarily consistent, disrupting direct 2D comparisons; on the other hand, reliable detection of damage progression directly in 3D remains inherently more complex due to the technical hurdles of 3D change analysis itself—such as the precise registration of volumetric data or the quantification of subtle geometric variations [14] .

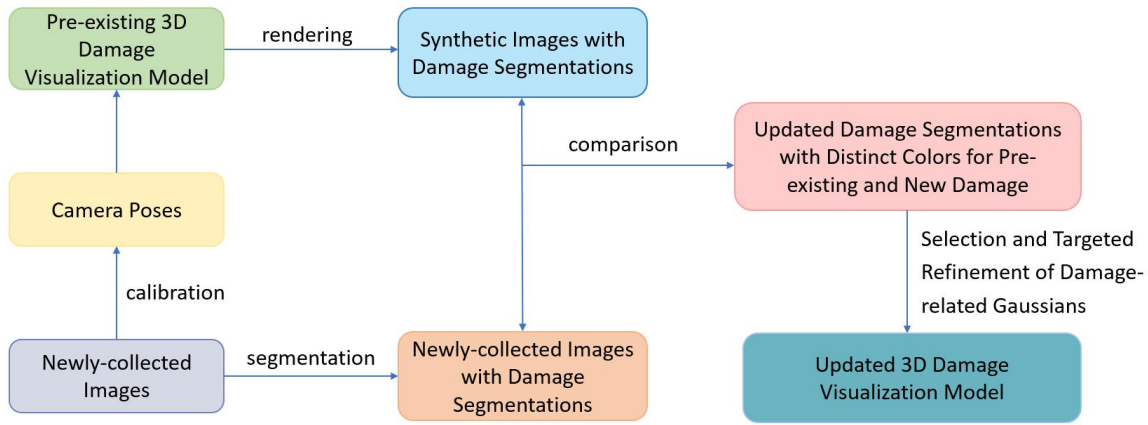

Figure 4. Procedure of GS-enabled digital twin updating for damage progression visualization

Once new damage is identified, updated masks are generated with distinct colors for pre-existing and newly detected damage. These masks are subsequently used to select damage-related Gaussians and facilitate their targeted refinement, following a procedure analogous to that described in the hierarchical 3D damage visualization workflow (Section 2.3). This localized update obviates the need for computationally intensive full re-reconstruction, enabling the digital twin to evolve easily and frequently in tandem with the physical structure. The result is a periodically synchronized model that provides stakeholders with a continuously accurate representation of structural health, supporting proactive maintenance strategies for infrastructure.

*2.5 Summary*

This study proposes adapting GS as a digital twin solution for 3D damage visualization of civil structures. Rather than overhauling GS's core pipeline, targeted modifications are introduced to address critical limitations that prevent the original GS framework from effectively supporting damage visualization tasks. Specifically, key contributions include: replacing raw input images with damage-segmented images as the optimization objective, enabling a shift from the original GS's appearance-inclusive but semantically agnostic scene representation to one with explicit semantic awareness of damage; developing a hierarchical training strategy that reduces computational costs and controls model size, signifying a clear advancement over the traditional GS approach's undifferentiated processing of all image data, which wastes resources on non-critical regions; and designing a dedicated mechanism for damage progression tracking and digital twin updating—a functionality entirely absent from the standard GS pipeline, which is inherently limited to static scene reconstruction. Together, these adaptations transform GS from a general-purpose view synthesis tool into a specialized digital twin framework tailored for the unique demands of civil infrastructure 3D damage visualization and structural health monitoring.

The advantages of the proposed GS-enabled method are multifaceted, as follows:

Unlike photogrammetric 3D reconstruction— which relies heavily on feature extraction and dense point matching— Gaussian Splatting (GS) leverages a parametric model training strategy for reconstruction. This key distinction endows GS with a notable advantage in handling image regions featuring repetitive textures or no distinct texture, a scenario where photogrammetric methods often struggle. This advantage further enables the direct integration of damage visualization into the GS-based 3D reconstruction workflow, even when damage-segmented images exhibit undifferentiated pixel values across color-coded damage regions. Specifically, the one-step 3D damage visualization facilitated by GS addresses two critical limitations of traditional approaches: it circumvents both the possible segmentation errors and stringent camera pose precision requirements inherent to single-view damage casting, while simultaneously ensuring multi-view consistency. In turn, this consistency delivers a more robust and error-tolerant representation of damage within the structural context of civil infrastructure.

Furthermore, Gaussian Splatting (GS) adopts an explicit, discrete scene representation built from Gaussian primitives—endowing it with controllability over regional adjustments, a capability far more pronounced when compared with the implicit continuous scene representation in NeRF-based reconstructions. This controllability enables a hierarchical reconstruction paradigm: low-resolution images first establish the overall structural geometry, while high-resolution damage-segmented images are solely deployed to refine damage-associated Gaussian primitives, thus achieving detailed damage representation. By confining high-fidelity processing exclusively to damage-relevant regions, this paradigm improves the utilization of computing resources.

GS also exhibits remarkable novel-view synthesis capabilities, which facilitate direct 2D cross-temporal comparisons of damage captured in different inspection surveys. This not only circumvents the challenges of 3D-based damage progression tracking—given that 3D change detection itself remains complex, considering precise volumetric registration and the quantification of subtle geometric variations—but also allows the GS-enabled model to better fulfill its role as a digital twin. Specifically, it enables convenient and frequent synchronization of the digital model with the physical structure's actual damage state.

## 3. Demonstration

In this section, the proposed methodology—comprising GS-enabled 3D damage visualization to mitigate single-view segmentation errors, a hierarchical reconstruction strategy for computational efficiency, and convenient digital twin updates using newly acquired inspection data—will be demonstrated using an open-source synthetic dataset tailored for post-earthquake structural inspections, referred to as the QuakeCity dataset. Developed using the physics-based graphics models (PBGM) proposed by Hoskere et al. [40] [41] , QuakeCity integrates surface damage textures derived from global and local finite element analysis (FEA) of building structures to generate photo-realistic structural models. Images are rendered from simulated multi-view UAV surveys of damaged buildings within a city environment, with each image accompanied by six types of annotations—three damage masks (cracks, spalling, exposed rebar), component labels, component damage states, and a depth map—as illustrated in Figure 5.

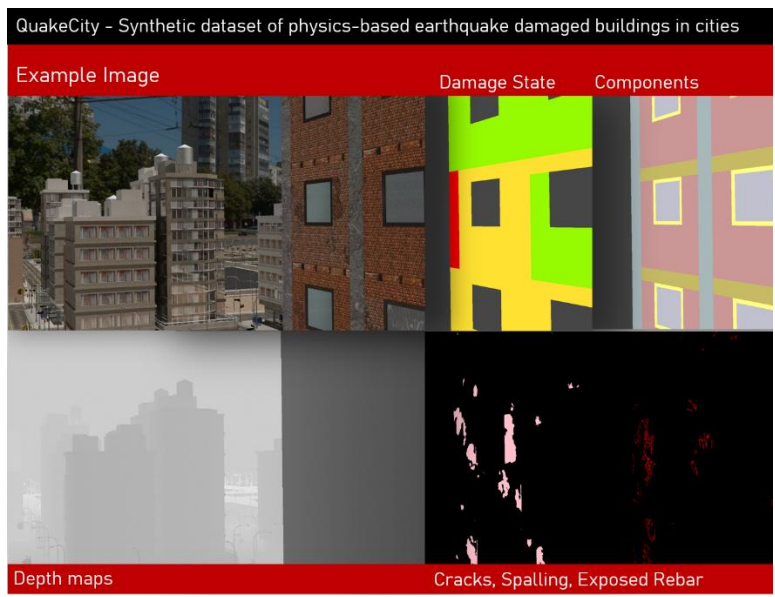

Figure 5. Examples of the QuakeCity dataset

Sixty high-resolution images from the QuakeCity dataset capturing the first three floors of a building, along with corresponding damage masks for spalling and cracks, are used to demonstrate the proposed method. Both the images and masks have a resolution of 1920×1080. These images are processed using the COncurrent Localization and Mapping (COLMAP) [42] [43] pipeline to perform SfM, yielding camera intrinsics, poses (positions and orientations), and a sparse 3D point cloud for subsequent use. The same set of images is then input into the standard GS pipeline, generating a 3D reconstruction model visualized in Figure 6, which provides a detailed and realistic representation of the scene.

Note that the background artifacts stem from insufficient view coverage and unsuccessful reconstruction of background objects. These artifacts can be easily avoided via two straightforward approaches: one is to define the scope of the target building and remove all the extraneous Gaussian primitives outside this boundary post-reconstruction; the other is to mask out irrelevant background regions in the 2D images prior to initiating the 3D reconstruction process. Similarly, artifacts observed on the upper portion of the reconstructed model result from inadequate view coverage of the building's upper floors—while only the bottom three floors were designated as the target, some of the 60 images nonetheless capture partial (yet insufficient) data of the upper floors. Such artifacts would be avoided with a comprehensive inspection survey that ensures full view coverage of the entire building.

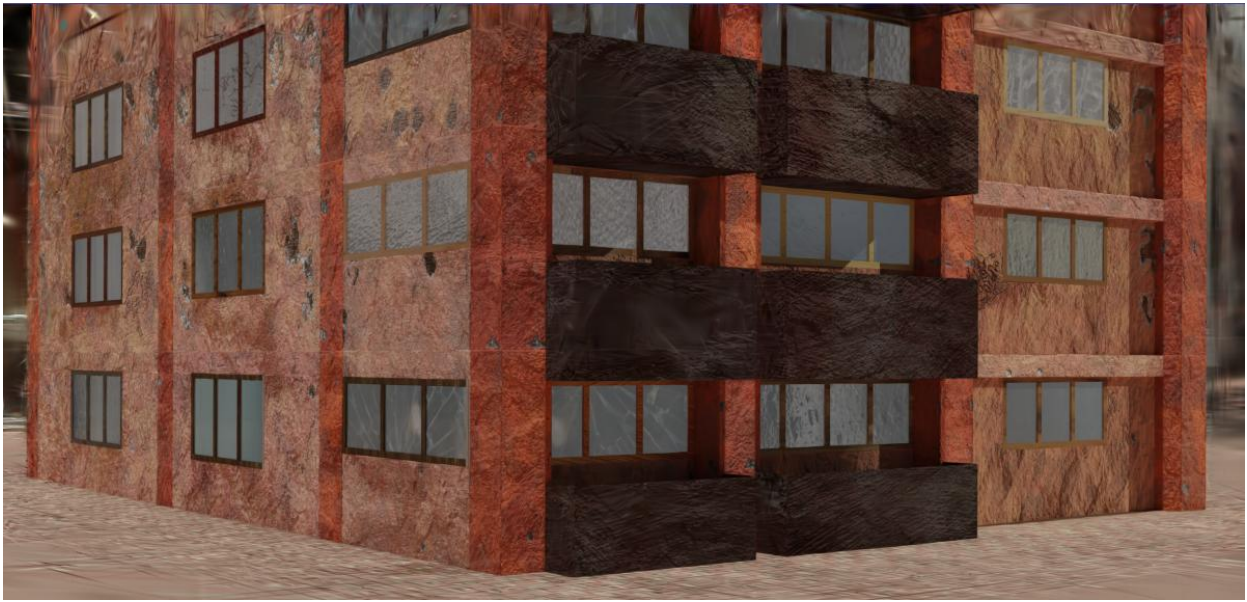

Figure 6. GS-based 3D reconstruction results using 60 example images from the QuakeCity dataset

### 3.1 Gaussian Splatting-enabled 3D damage visualization

To demonstrate GS's capability for 3D damage visualization, spalling masks are overlaid on the original images (see examples in Figure 7) to generate damage-segmented images. These segmented images are then fed into the 3D damage visualization workflow described in Section 2.2, which integrates damage segmentation results into the GS optimization process. The resulting 3D damage visualization model, shown in Figure 8, intuitively visualizes spalling damage within its spatial context, embedding 2D segmentation masks into a coherent 3D structural framework for enhanced interpretability.

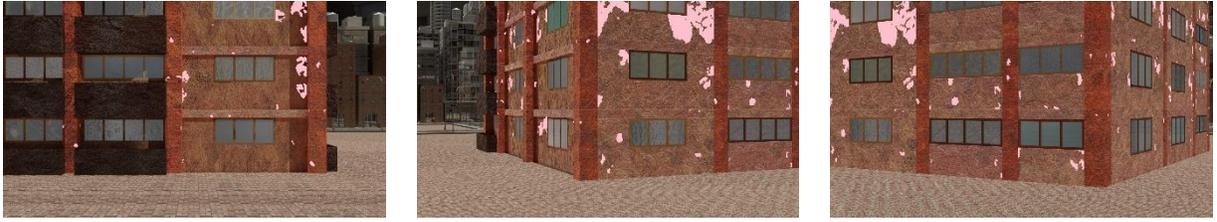

Figure 7. Examples of spalling-segmented images

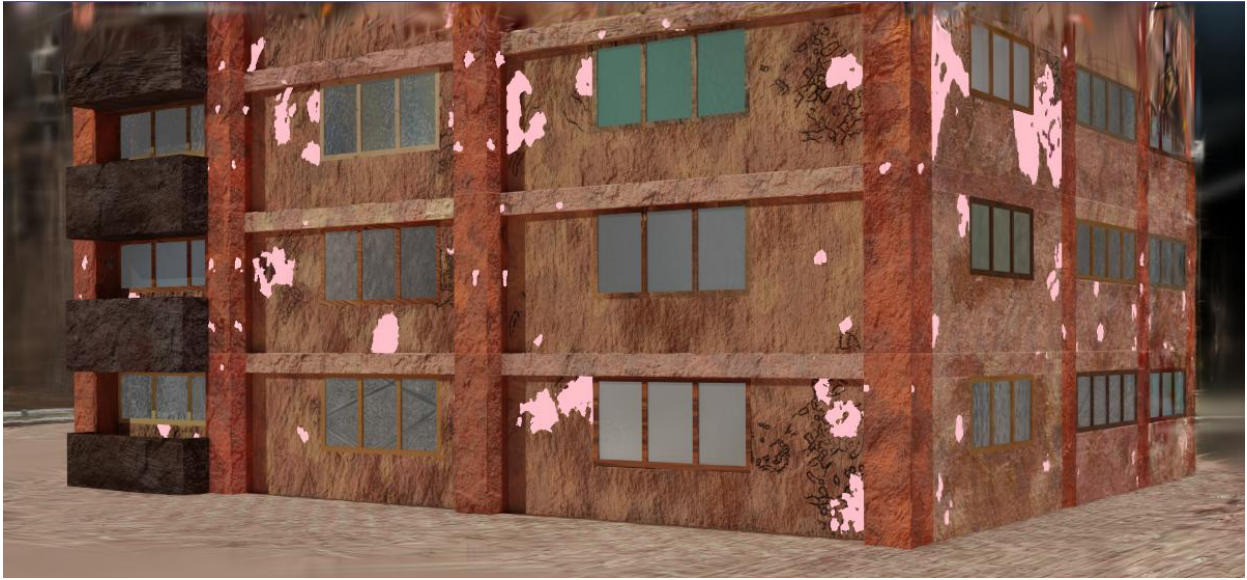

Figure 8. GS-enabled 3D damage visualization results using 60 spalling-segmented images

Furthermore, to demonstrate the superiority of GS over photogrammetric 3D reconstruction in handling smooth, featureless regions—such as color-coded damage segmentations—an extreme scenario is considered here. As shown in Figure 9, each structural component (wall, beam, column, window frame, window panel, balcony, and slab) is assigned to a distinct color. Using the same set of camera poses estimated by COLMAP from the original images, these component-segmented images are then used for 3D reconstruction via both the photogrammetric software Agisoft Metashape [21] and the GS-based pipeline. As illustrated in Figure 10, photogrammetric 3D reconstruction fails to yield valid results due to the lack of dense one-to-one matching features—a limitation directly stemming from undifferentiated pixel values across the image; in contrast, GS-based 3D reconstruction successfully captures the structural component segmentations. This notable advantage in handling repetitive or textureless image regions arises because GS leverages parametric model training to learn scene representations, rather than relying solely on feature extraction and dense point matching for 3D reconstruction.

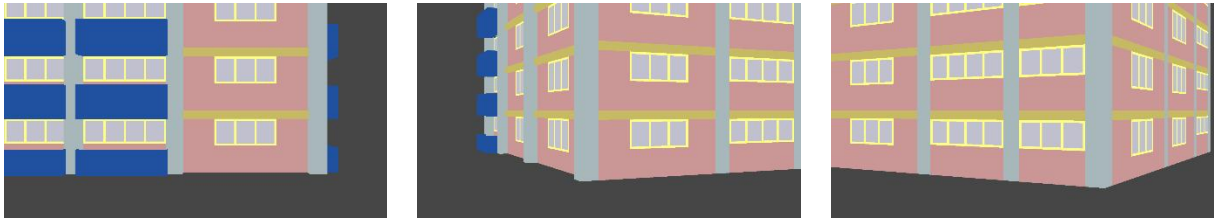

Figure 9. Examples of component-segmented images

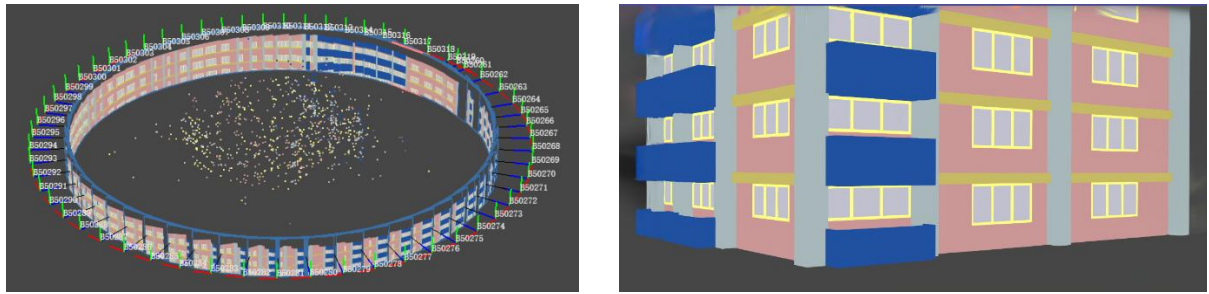

(a) photogrammetric reconstruction            (b) GS-based reconstruction

Figure 10. 3D reconstruction results using 60 spalling-segmented images

Additionally, to verify GS's capability to mitigate segmentation errors in damage visualization, intentional errors are introduced into ground-truth spalling segmentations to create contradictions between multi-view segmentations. These errors include random addition (green circle) or removal (yellow circles) of irregular regions in spalling masks, as illustrated in Figure 11. The resulting segmented images with conflicting annotations are then processed through the GS-enabled 3D damage visualization workflow, generating the 3D damage visualization model shown in Figure 12. Figure 13 presents the spalling-segmented image rendered from the 3D damage visualization model shown in Figure 12, viewed from the same perspective as those in Figure 11. Notably, most of the intentionally introduced segmentation errors have been effectively mitigated, resulting in rendered segmentations that closely align with the ground truth. The results validate that GS's inherent multi-view consistency optimization effectively suppresses segmentation inconsistencies, yielding a consistent 3D damage visualization (Figure 12) aligned with the model from accurate masks (Figure 8).

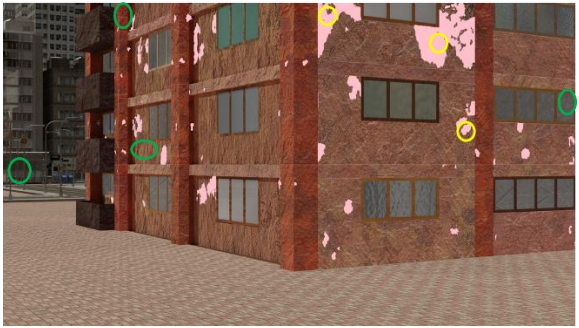 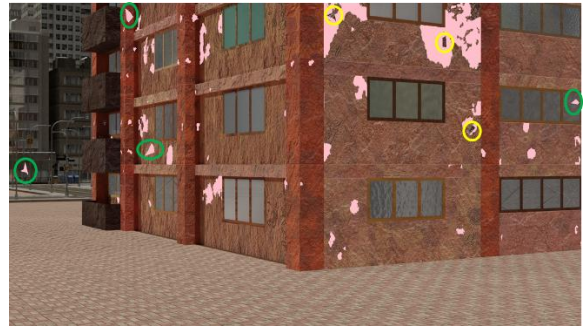

(a) ground-truth spalling segmentations          (b) modified spalling segmentations

Figure 11. An example of intentionally introduced errors in spalling segmentations

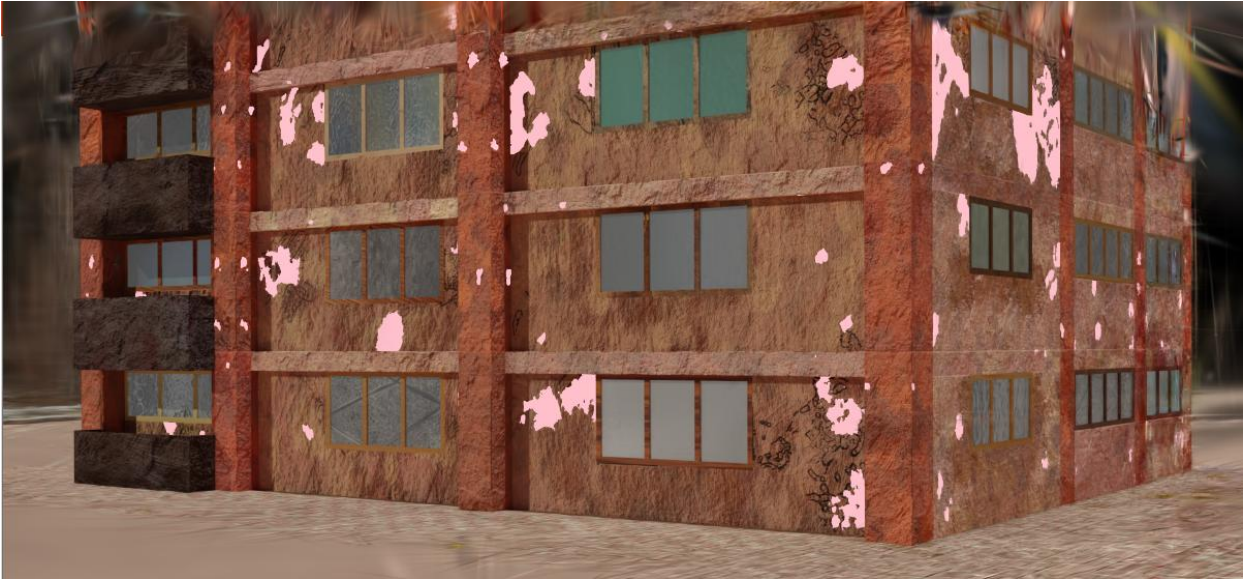

Figure 12. GS-enabled 3D damage visualization results using 60 spalling-segmented images with segmentation errors

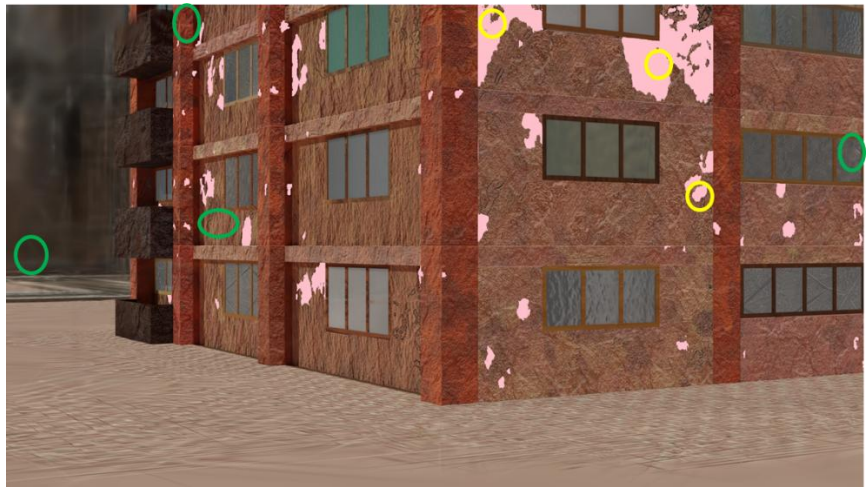

Figure 13. A spalling-segmented image rendered from the 3D GS model presented in Figure 12

*3.2 Hierarchical 3D damage visualization via Gaussian Splatting*

To demonstrate the hierarchical 3D damage visualization strategy, crack masks are overlaid on original images to generate damage-segmented images. Pre-processing steps are performed to facilitate the demonstration: instead of including all damage across the building's four facades, only a single crack segment is retained, resulting in 12 images that capture the reserved crack damage. Additionally, the red color of crack masks in the QuakeCity dataset is changed to blue to enhance contrast with the building's red-toned façade (see examples in Figure 14), facilitating the selection of Gaussians related to crack damage during the damage visualization process.

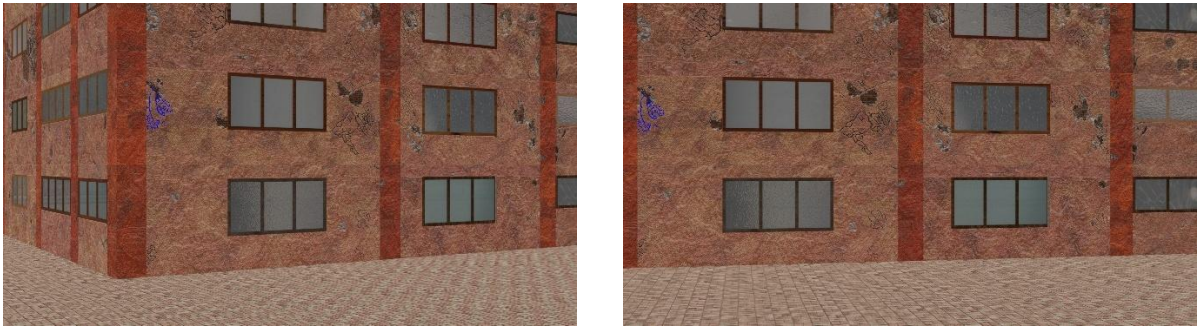

Figure 14. Examples of crack-segmented images

The hierarchical 3D damage visualization procedure begins with down sampling the crack-segmented images to a low resolution of 480×270. Using the camera poses derived from the original-resolution images and taking the low-resolution crack-segmented images as the ground truth for the loss function, a baseline damage visualization model shown in Figure 15 is generated. The reconstruction lacks fine details when compared to the models in Figure 6 or Figure 8, which were generated using high-resolution images. However, the model provides sufficient structural context, with crack regions visualized within it, enabling the selection of crack-related Gaussians for further refinement using high-resolution images.

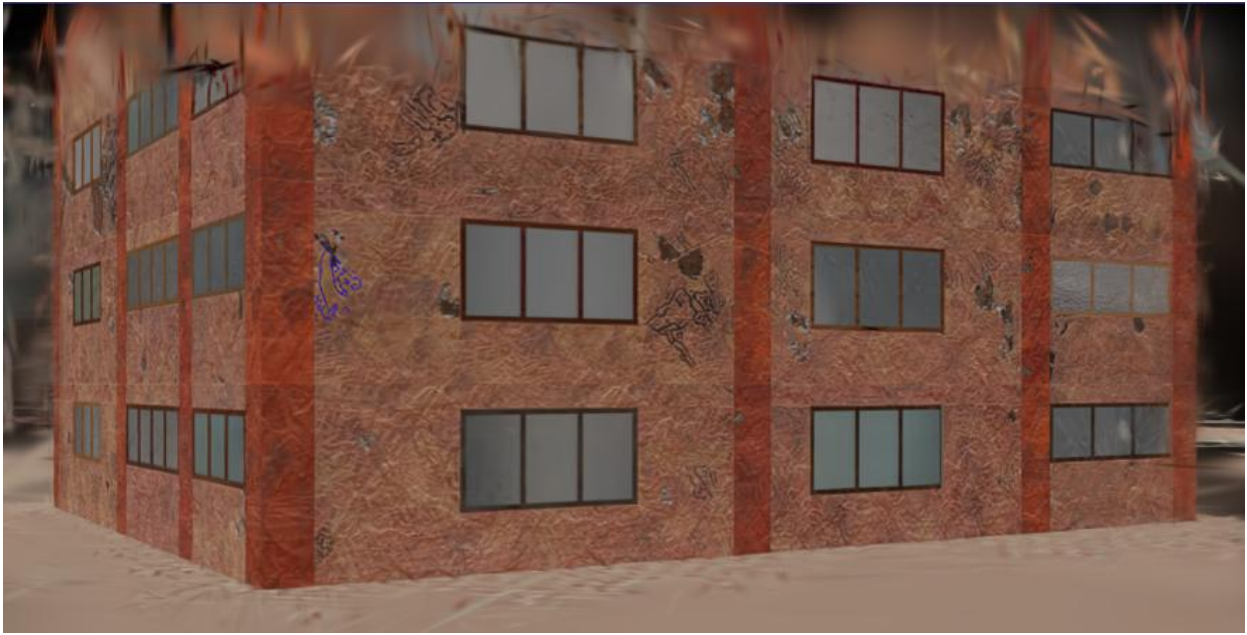

Figure 15. GS-enabled 3D damage visualization results using 60 low-resolution crack-segmented images

Next, Gaussian primitives associated with cracks are isolated from others through color filtering, leveraging the distinct blue tone assigned to cracks to differentiate them from the structural context. To mitigate potential errors arising from exclusive reliance on color-based selection, the color-filtered 3D Gaussians are projected onto each segmentation mask to validate their true association with the crack. Neighboring Gaussian primitives are then incorporated by searching within an adaptive radius, which will be expanded when necessary to ensure a sufficient number of neighbors are detected for capturing contextual structural details of the crack. The selected Gaussian primitives—visualized in Figure 16 (a)— are those that will undergo adjustment in the subsequent refinement phase. Additionally, the 2D projections of these primitives onto each image plane are computed, and a convex hull is generated to encompass the entire projection region; this hull is further slightly dilated to create a small buffer zone around the damage region, as illustrated in Figure 16 (b). In the subsequent refinement phase, the loss function between each target image and its corresponding rendered image is computed exclusively within this new mask, thereby confining adjustments to the damage regions and preventing unintended impacts on the rest of the structure from modifications to damage-related Gaussians.

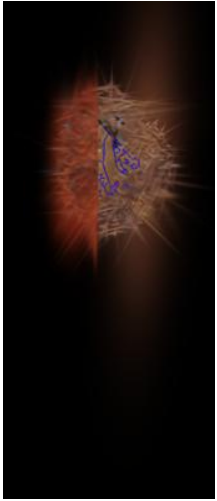 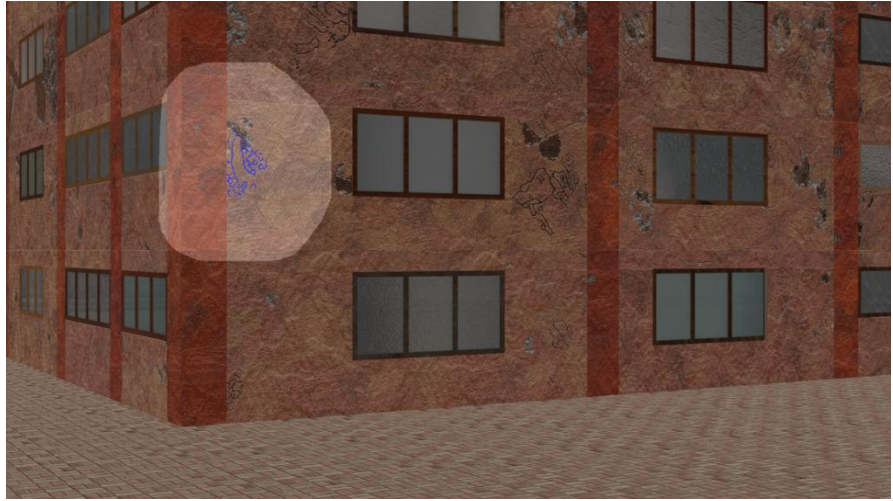

(a) selected Gaussians            (b) convex hull mask (overlaid on the corresponding crack-segmented image)

Figure 16. Results of preprocessing steps prior to refinement of the GS-enabled 3D damage visualization model

Subsequently, the selected Gaussian primitives are fine-tuned or retrained using eight high-resolution images with crack segmentation masks—chosen from the total of 12 images that capture the crack—for their optimal angles in depicting the targeted crack damage. As outlined in Section 2.3, two strategies are employed: fine-tuning and retraining. The former adjusts only parameters related to position, color, opacity, scaling, and rotation without altering the number of Gaussian primitives, while the latter retains GS's densification mechanisms, including cloning, splitting, and pruning—to modify primitive quantities. Figure 17 presents a comparative example of the crack segmentations rendered from the refined GS model via each strategy, set against the corresponding ground-truth image—which is one of the multi-view crack-segmented images used for the GS-enabled 3D reconstruction. Both strategies significantly enhance crack representation: the retraining strategy enables superior recovery of crack details, whereas the fine-tuning strategy tends to produce slightly wider crack depictions and omit certain discontinuities—limitations arising from its inability to adjust the number of Gaussian primitives.

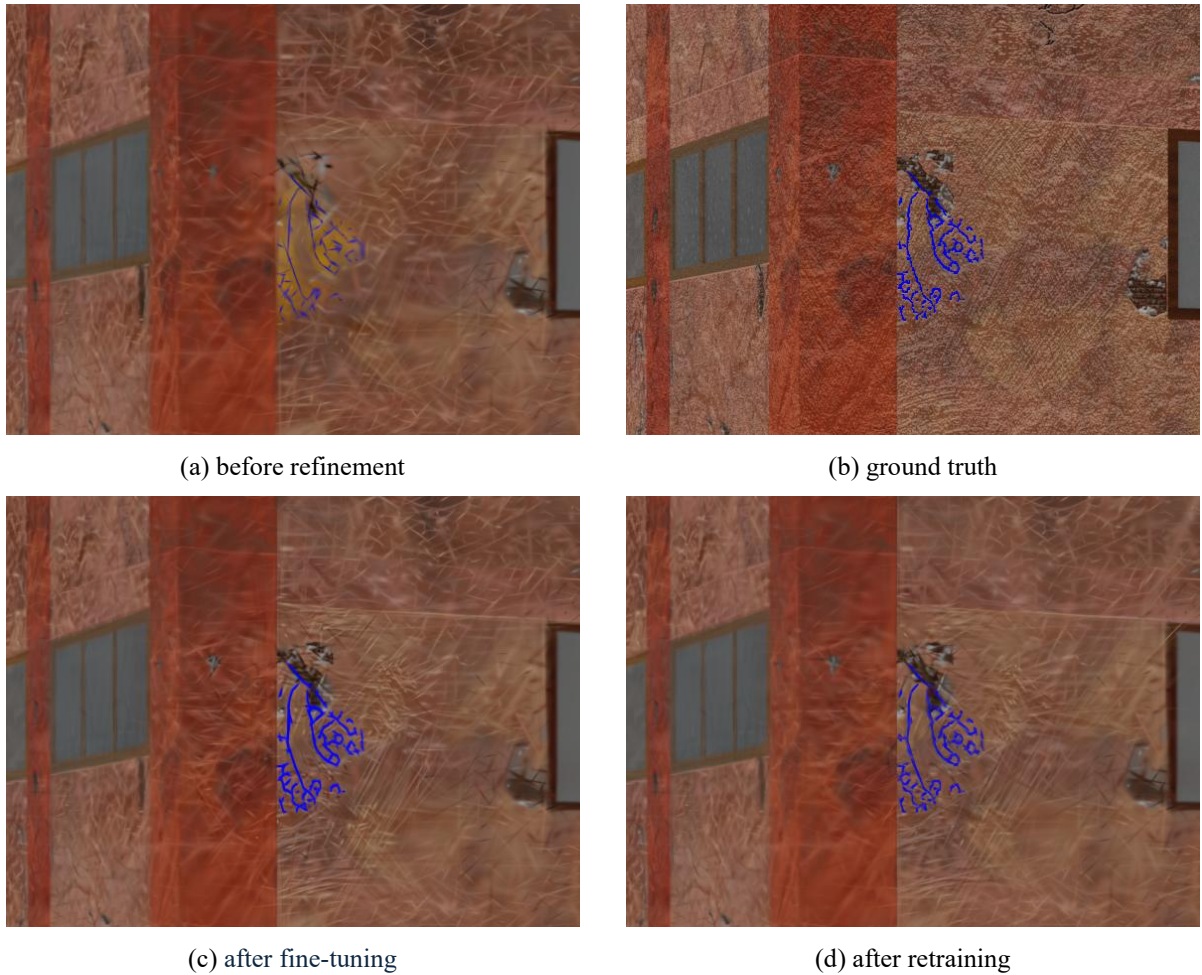

(a) before refinement

(b) ground truth

(c) after fine-tuning

(d) after retraining

Figure 17. Crack-segmented images rendered from the 3D GS model before and after refinement

Regarding efficiency, runtime experiments for the two reconstruction approaches—reconstruction of the entire structure using high-resolution damage-segmented images, and the proposed hierarchical training strategy—were conducted on a device equipped with a single NVIDIA RTX 4090 graphics card. For reconstruction with 60 high-resolution (1920×1080) images: training for 7,000 epochs took 4 minutes and 15 seconds, yielding unsatisfactory results; in contrast, training for 30,000 epochs required 22 minutes and 42 seconds, which produced high-quality reconstruction outcomes for the entire structure. By comparison, reconstruction with 60 low-resolution (480×270) images took only 59 seconds to complete 7,000 epochs—yet this was sufficient to generate reconstruction results of acceptable quality (see Figure 15) for both the overall structural context and the subsequent selection of crack-related Gaussian primitives. This efficiency gain stems from the reduced detail in low-resolution images, which minimizes training computational demands. Preprocessing steps including the selection of crack-related Gaussians and the generation of convex hull masks, required only a few seconds. For the refinement stage of the hierarchical strategy: fine-tuning crack-related Gaussians for 1,000 iterations (to obtain the result in Figure 17(c)) took 57 seconds, while retraining the same set of Gaussians for 1,000 iterations (for Figure 17(d)) took 6 minutes and 54 seconds— with the extra time primarily attributed to the densification step, which was executed every 100 iterations. Even the relatively time-consuming variant of the hierarchical strategy—where damage-related Gaussians are retrained—reduces the total reconstruction time by over 60% compared to full reconstruction of the entire structure using high-resolution images. These runtime experiments confirm that the proposed hierarchical training strategy effectively reduces overall reconstruction time while maintaining critical quality for structural damage visualization.

*3.3 Gaussian Splatting-enabled digital twin updating for damage progression visualization*

To demonstrate the tracking of damage progression in digital twins via GS, an additional eight images from the QuakeCity dataset— depicting a new crack damage instance alongside the existing one—are employed (see Figure 18). First, these eight images are incorporated into the COLMAP pipeline alongside the existing 60 images to estimate their camera poses within the coordinate system of the existing GS model. Next, images corresponding to these eight perspectives are rendered from the existing damage visualization model (containing a single crack segment), with an example shown in Figure 19 (a). A comparison with the segmentation results of the corresponding newly collected image (as shown in Figure 18 (left)) highlights regions where new damage has emerged, resulting in an updated damage mask: pre-existing cracks are colored blue, and newly detected cracks are colored green (Figure 19 (b)).

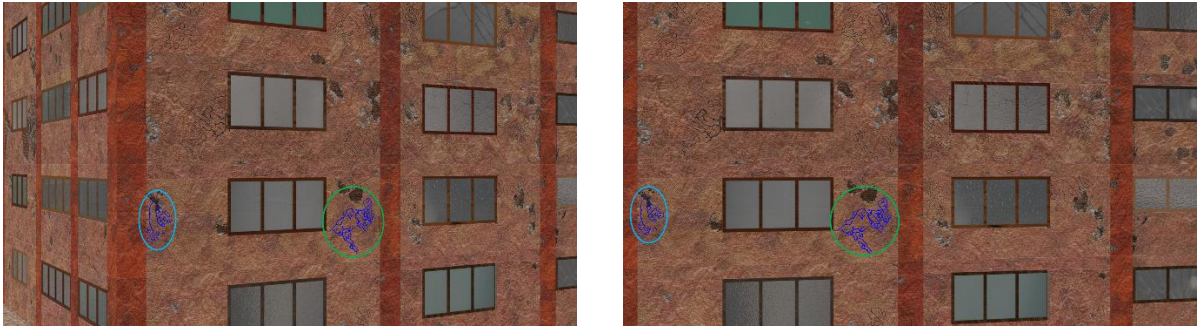

Figure 18. Examples of eight new images (two shown here) depicting a new crack (marked with a green circle) alongside the existing crack (marked with a blue circle)

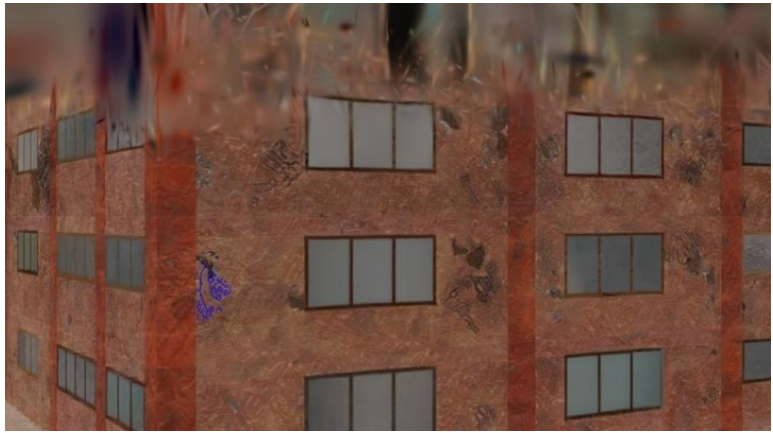

(a) crack-segmented image rendered from the existing damage visualization model (same perspective as Figure 18 left)

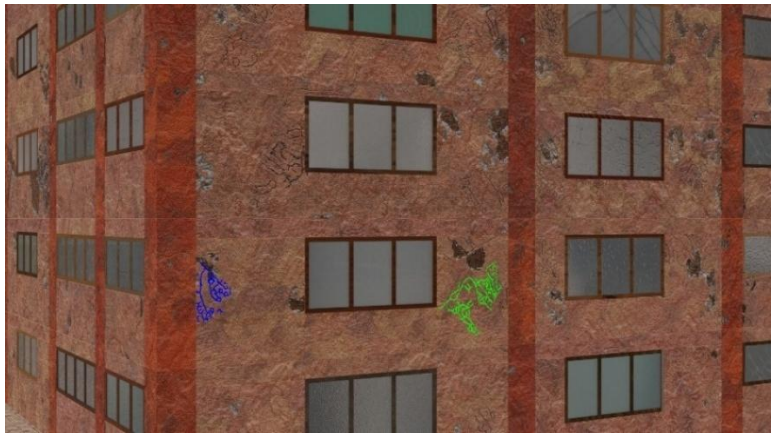

(b) updated crack-segmented image

Figure 19. Tracking of damage progression enabled by the novel view synthesis capacity of GS

These updated masks are then used to select Gaussian primitives related to both pre-existing and newly emerged damage. Subsequently, targeted refinement of these primitives is conducted following the fine-tuning or retraining procedures outlined in the hierarchical 3D damage visualization workflow

(Section 2.3), yielding an updated 3D model (as shown in Figure 20) that visualizes the progression of damage over time. The results validate the proposed GS-enabled method's capability to effectively track damage progression in digital twins by distinguishing new damage from pre-existing damage and seamlessly incorporating new damage observations into the 3D visualization model.

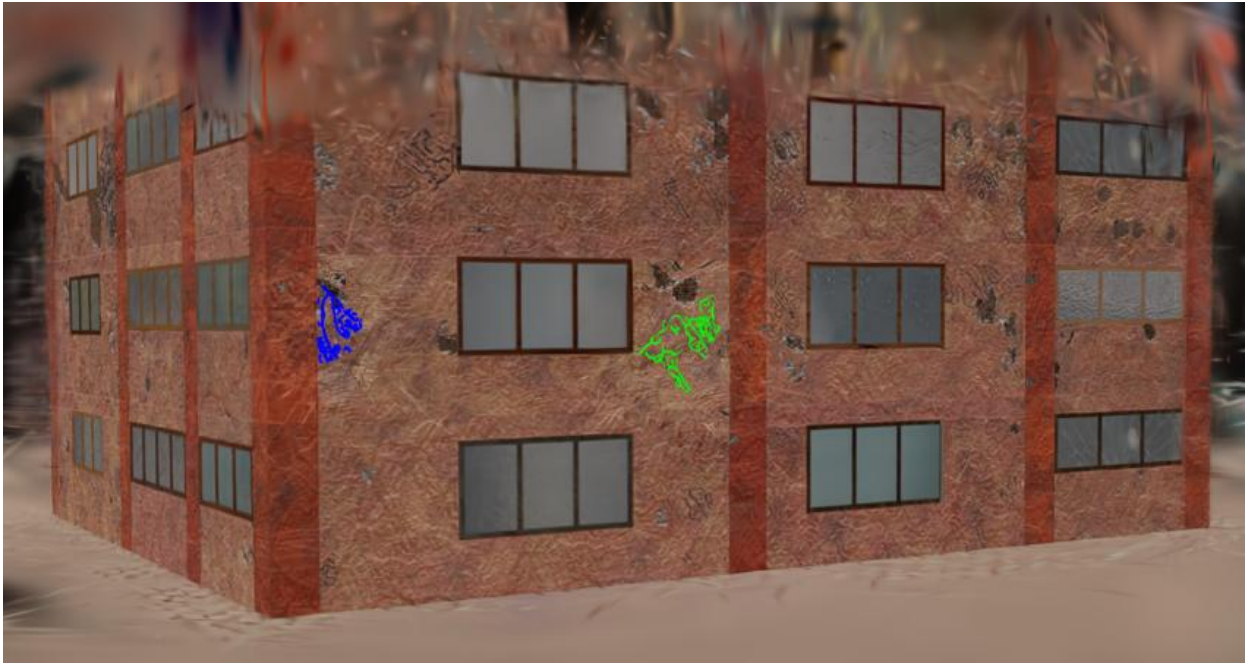

Figure 20. Updated 3D damage visualization model depicting damage progression

## 4. Conclusions

This research proposed a novel Gaussian Splatting-empowered digital twin method for 3D damage visualization of civil structures. The effectiveness of this method was demonstrated on an open-source synthetic image dataset for post-earthquake inspections. Results showed that the method accurately visualized 3D damage by integrating 2D damage segmentation masks into the 3D reconstruction loss function and leveraging multi-view consistency to mitigate single-view segmentation errors. Additionally, a hierarchical reconstruction strategy was employed to conserve computational resources, first generating a coarse-grained model from low-resolution images, then applying targeted fine-tuning or retraining with high-resolution data to capture critical damage details. The approach also enabled the identification of emerging damage by comparing damage segmentations from newly collected images taken from arbitrary views with those rendered from the pre-existing 3D GS model. Seamless updates to the digital twin were then achieved by utilizing updated damage segmentations—incorporating distinct colors for existing and emerging damage—to refine damage details of the reconstruction model, thus obviating the need for full re-reconstruction with each inspection survey.

Overall, the method achieved a balance between computational efficiency and damage fidelity, providing a robust and scalable solution for 3D damage visualization in civil infrastructure digital twins. This work laid the foundation for advanced structural health monitoring and maintenance strategies by conveniently synchronizing digital models with real-world damage progression. For future work, the approach will be validated on real buildings to address potential issues that may arise, and a user-friendly graphical interface (GUI) will be developed to facilitate the convenient utilization of each function. Additionally, developing automated damage severity assessment frameworks within the digital twin workflow would enable proactive maintenance decisions, bringing the approach closer to practical deployment in smart infrastructure systems.

## Acknowledgements

The authors gratefully acknowledge the support provided by the National Natural Science Foundation of China (NSFC) under Grant Nos. 52325802 and 52361165658. The authors would also like to acknowledge the support of Dr. Vedhus Hoskere, Assistant Professor from the University of Houston, for providing access to the QuakeCity dataset, as well as Dr. Junhwa Lee, Assistant Professor from Pukyong National University, for the inspirational discussions that enriched this work.